\pdfoutput=1

\documentclass[11pt]{article}

\usepackage{EMNLP2023}

\usepackage{times}
\usepackage{latexsym}

\usepackage[T1]{fontenc}

\usepackage[utf8]{inputenc}

\usepackage{microtype}

\usepackage{inconsolata}
\usepackage{enumitem}
\usepackage{graphicx}
\usepackage{array}
\usepackage{amssymb}
\usepackage{pifont}
\newcommand{\cmark}{\ding{51}}%
\newcommand{\xmark}{\ding{55}}%

\usepackage{amsmath}
\usepackage{makecell}
\usepackage{multirow}

\newcommand{\ignore}[1]{}

\makeatletter
\DeclareRobustCommand\onedot{\futurelet\@let@token\@onedot}
\def\@onedot{\ifx\@let@token.\else.\null\fi\xspace}
\makeatother

\definecolor{MyDarkBlue}{rgb}{0,0.08,1}
\definecolor{MyDarkGreen}{rgb}{0.02,0.6,0.02}
\definecolor{MyDarkRed}{rgb}{0.8,0.02,0.02}
\definecolor{MyDarkOrange}{rgb}{0.40,0.2,0.02}
\definecolor{MyPurple}{RGB}{111,0,255}
\definecolor{MyRed}{rgb}{1.0,0.0,0.0}
\definecolor{MyGold}{rgb}{0.75,0.6,0.12}
\definecolor{MyDarkgray}{rgb}{0.66, 0.66, 0.66}

\newif\iftodo

\todofalse

\newcommand\MyDarkBluesout{\bgroup\markoverwith{\textcolor{MyDarkBlue}{\rule[0.5ex]{2pt}{1.2pt}}}\ULon}

\title{MARRS: Multimodal Reference Resolution System}

\author{Halim Cagri Ates, Shruti Bhargava${}^1$, Site Li, Jiarui Lu, Siddhardha Maddula, \\
        \textbf{Joel Ruben Antony Moniz${}^2$, Anil Kumar Nalamalapu, Roman Hoang Nguyen, }\\
        \textbf{Melis Ozyildirim, Alkesh Patel, Dhivya Piraviperumal${}^3$, Vincent Renkens, }\\
        \textbf{Ankit Samal, Thy Tran, Bo-Hsiang Tseng${}^4$, Hong Yu${}^5$, Yuan Zhang, Rong Zou\thanks{\enspace Authors listed in alphabetical order}}\\
        \{${}^1$shruti\_bhargava, ${}^2$joelmoniz, ${}^3$dhivyaprp, ${}^4$bohsiang\_tseng, ${}^5$hong\_yu\}@apple.com \\
        Apple}

\begin{document}

\maketitle
\begin{abstract}

Successfully handling context is essential for any dialog understanding task. This context maybe be conversational (relying on previous user queries or system responses), visual (relying on what the user sees, for example, on their screen), or background (based on signals such as a ringing alarm or playing music). In this work, we present an overview of MARRS, or Multimodal Reference Resolution System, an on-device framework within a Natural Language Understanding system, responsible for handling conversational, visual and background context. In particular, we present different machine learning models to enable handing contextual queries; specifically, one to enable reference resolution, and one to handle context via query rewriting. We also describe how these models complement each other to form a unified, coherent, lightweight system that can understand context while preserving user privacy.

\end{abstract}

\section{Introduction}
\label{sec:intro}
Fast-paced advancements across modalities have presented exciting opportunities and daunting challenges for dialogue agents. The ability to seamlessly integrate and interpret different types of information is crucial to achieve human-like understanding. One fundamental aspect of dialogue agents, therefore, is their ability to understand references to context, which is essential to enable them carry out coherent conversations. Traditional reference resolution systems \cite{yang-etal-2019-end-end-neural} are not sufficient for multiple modalities in a dialogue agent.

In this work, we introduce MultimodAl Reference Resolution System (MARRS), targeted to understand and resolve diverse context understanding use cases. MARRS leverages multiple types of context to understand a request, while completely running on-device, keeping memory and privacy as key design factors. The key objective of MARRS is two-fold: first, to track and maintain coherence during multiple turns of a conversation, and second, to leverage visual context to enhance context understanding. It thus aims to provide a centralized domain agnostic solution to diverse discourse and referencing tasks including, but not limited to\footnote{Examples shown are author-created queries based on anonymized and randomly sampled virtual assistant logs.}:

\begin{figure}
  \centering
  \includegraphics[width=1.0\linewidth]{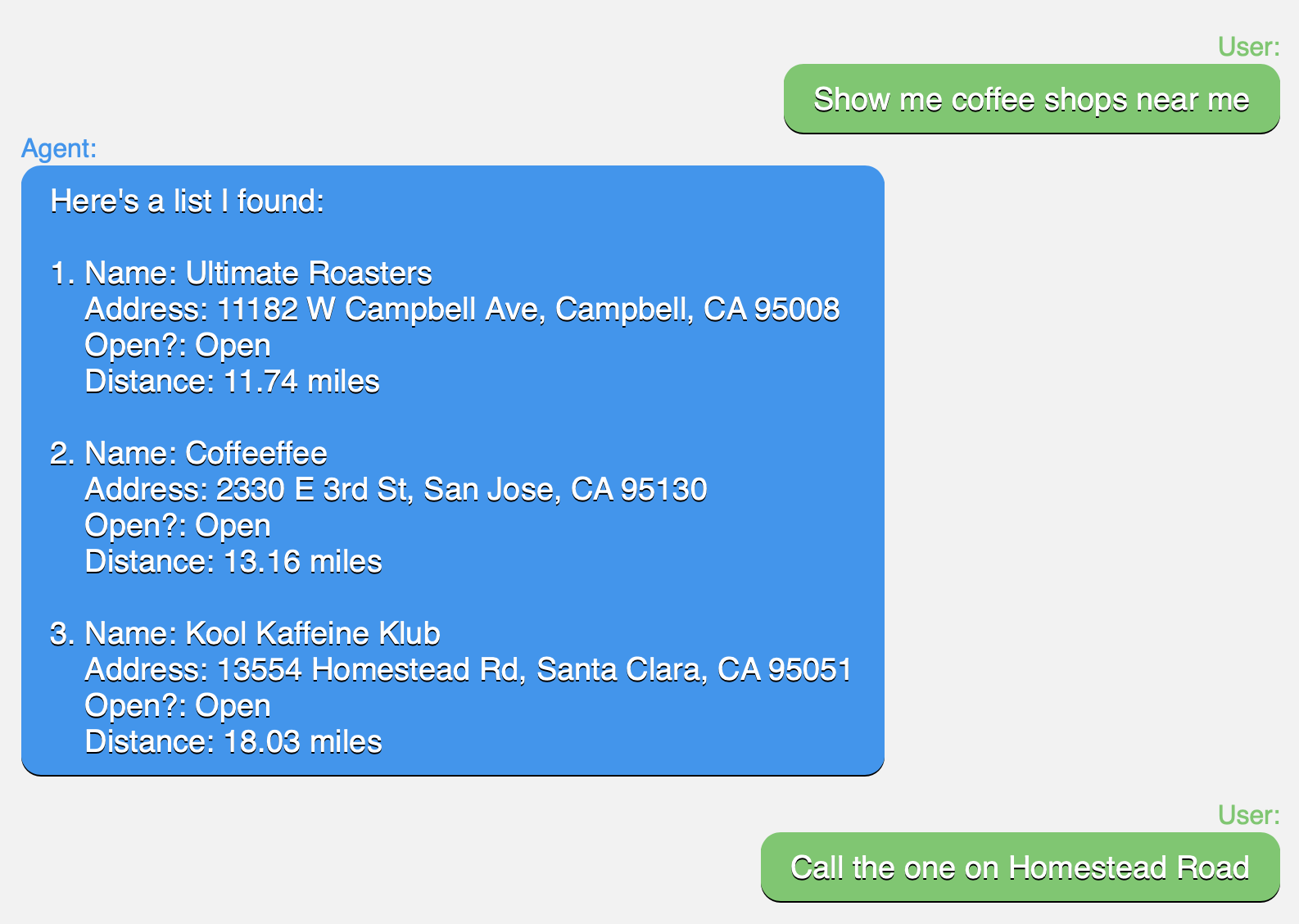}
  \caption{An example of Conversational Entity Resolution. All coffee shop names shown are author-created.}
  \label{fig:conv_rr_example_flow}
\end{figure}

\paragraph{Anaphora Resolution}
     \begin{verbatim}
User: What is Ohio's capital?
Agent: Columbus is the capital of Ohio.
User: How far away is it?
\end{verbatim}
\paragraph{Ellipsis Resolution}
\begin{verbatim}
User: What is the currency of France?
Agent: The Euro is the currency of France.
User: What about United States?
\end{verbatim}
\paragraph{Screen Entity Resolution}
\begin{verbatim}
User: Share this number with John.
\end{verbatim}
\paragraph{Conversational Entity Resolution} Note that here, the entity may be a part of the interaction without being explicitly mentioned. 
\begin{verbatim}
User: Show me pharmacies near me.
Agent: Here are some near you: <list>
User: Call the second one
\end{verbatim}
\paragraph{Background Entity Resolution}
\begin{verbatim}
*alarm starts ringing*
User: Switch it off
\end{verbatim}
\paragraph{Correction by Repetition}
\begin{verbatim}
User: What is the population of Australia
Agent: The population of Australia is ...
User: I meant Austria
\end{verbatim}

MARRS consumes the transcribed request, and as a part of the language understanding block, aims to allow for fluent conversations spanning multiple modalities. It takes screen entities, conversation history, and other contextual entities as input alongside the latest transcribed request; and outputs a context independent rewritten request as well as spans that link references to entities. In some cases, like \textit{conversational referencing} above, a span with entity id may be preferred by downstream components, while a rewritten query may be preferred in \textit{ellipsis resolution} for transparent low-effort adoption downstream. Central to the success of MARRS are its two components: the query rewriter and the reference resolver (comprising, in turn, of the mention detector and the mention resolver). The query rewriter aims to rewrite a user query to make it context independent, thereby making it self-contained. The mention detector and resolver on the other hand aim to generate reference spans.  

In this paper, we delve into the system design of MARRS, its components, the reasoning behind them and how they integrate together for efficient context understanding. Note that while we find our system highly efficient and performant, detailed benchmarks and results are outside the scope of this paper. We believe this work will foster an understanding of multimodal context understanding systems and pave the way for more sophisticated and contextually-aware agents.

\section{System Design}

The context carryover problem is usually tackled with coreference resolution \cite{ng2002improving}. Traditional coreference resolution systems often identify mentions and link the mention to entities in the previous context \cite{lee2017end}.  Another approach to address the problem is to rewrite the user request into a version which can be executed in a context independent way \cite{nguyen2021user, quan-etal-2019-gecor, 10.1145/3397271.3401323, Tseng2021CREADCR}. There are pros and cons for each of the two approaches. 

On one hand, coreference resolution provides spans with entities, which downstream systems can consume. This removes the need to perform entity linking again, which may add latency and/or errors, and also supports references to complex entities (like calendar events) where rewriting to a natural language query could be hard. On the other hand, the rewrite approach can handle not only the coreference resolution problem, but also other discourse phenomena such as intent carryover, corrections and disfluencies. Further, a coreference resolution system generates spans that need to be adopted by downstream systems, while a rewriting system reformulates the query itself, requiring no explicit adoption. In MARRS, we generate both reference spans and and query rewrites in order to take advantage of both approaches. See Figure~\ref{fig:marrs} for the design of MARRS.

\begin{table*}[!ht]
\begin{center}
\begin{tabular}{||m{8em} m{5em} m{5em} m{5em} m{5em} m{6em} ||} 
 \hline
 Previous work on Reference Resolution & Anaphora & Ellipses & Correction by Repetition & Screen Entity Resolution  & Conversational Entity Resolution\\ [0.5ex] 
 \hline\hline
  \citet{bohnet2023coreference} & \cmark & \xmark & \xmark & \xmark & \xmark  \\ 
 \hline
\citet{bhargava-etal-2023-referring} & \xmark & \xmark & \xmark & \cmark & \xmark\\
 \hline
 \citet{nguyen2021user} & \xmark & \xmark & \cmark & \xmark &  \xmark \\
 \hline
\citet{Tseng2021CREADCR}& \cmark & \cmark & \xmark & \xmark &  \xmark \\
 \hline
 \hline
\end{tabular}\
\end{center}
\caption{Comparison of previous work on reference resolution covering various use cases}
\label{table:1}
\end{table*}

There have been multiple prior works as shown in Table \ref{table:1}, trying to solve different aspects of reference resolution. A real-world dialogue system, however, requires the ability to simultaneous handle all of these aspects. In the MARRS system we do this using 2 major components, the Query Rewriter and the Reference Resolution System. The query rewrite component rewrites the current utterance with previous context, solving problems like anaphora and ellipses. Our reference resolution (or MDMR) component takes in contextual and screen entities and decorates the current utterance with entity information. This helps solve use cases related to screen, background and conversational references. Note that both the Query Rewriter and the Reference Resolution System are independent of each other; consequently, for efficency, they can be run in parallel. Overall, this system consumes dialog context, the current utterance, and entities as input, and produces a rewritten utterance and reference spans as output. Furthermore, the system has been designed to run on the (relatively low-power) device to preserve the privacy of the users.

Within our coreference resolution system, our system runs on \emph{all} user queries, since we do not know a priori if a user query requires resolution. Consequently, while end-to-end approaches have been proposed \cite{lee2017end}, we find it extremely beneficial for system performance to have a 2-stage pipeline: a light-weight Mention Detector (MD), followed by a more expensive Mention Resolver (MR) which is only run if MD detects a mention. We discuss each component in the following sections and detailed model architectures are provided in Section \ref{sec:model_diag}.

\subsection{Mention Detector}
The Mention Detector (MD) identifies sub-strings in the user utterance that can be grounded to one or more contextual entity. These are also known as referring expressions or mentions. Some examples of referring expressions are:
\begin{verbatim}
How big is [this house]
Where does [he] live
\end{verbatim}

\begin{figure}
  \centering
  \includegraphics[width=1.0\linewidth]{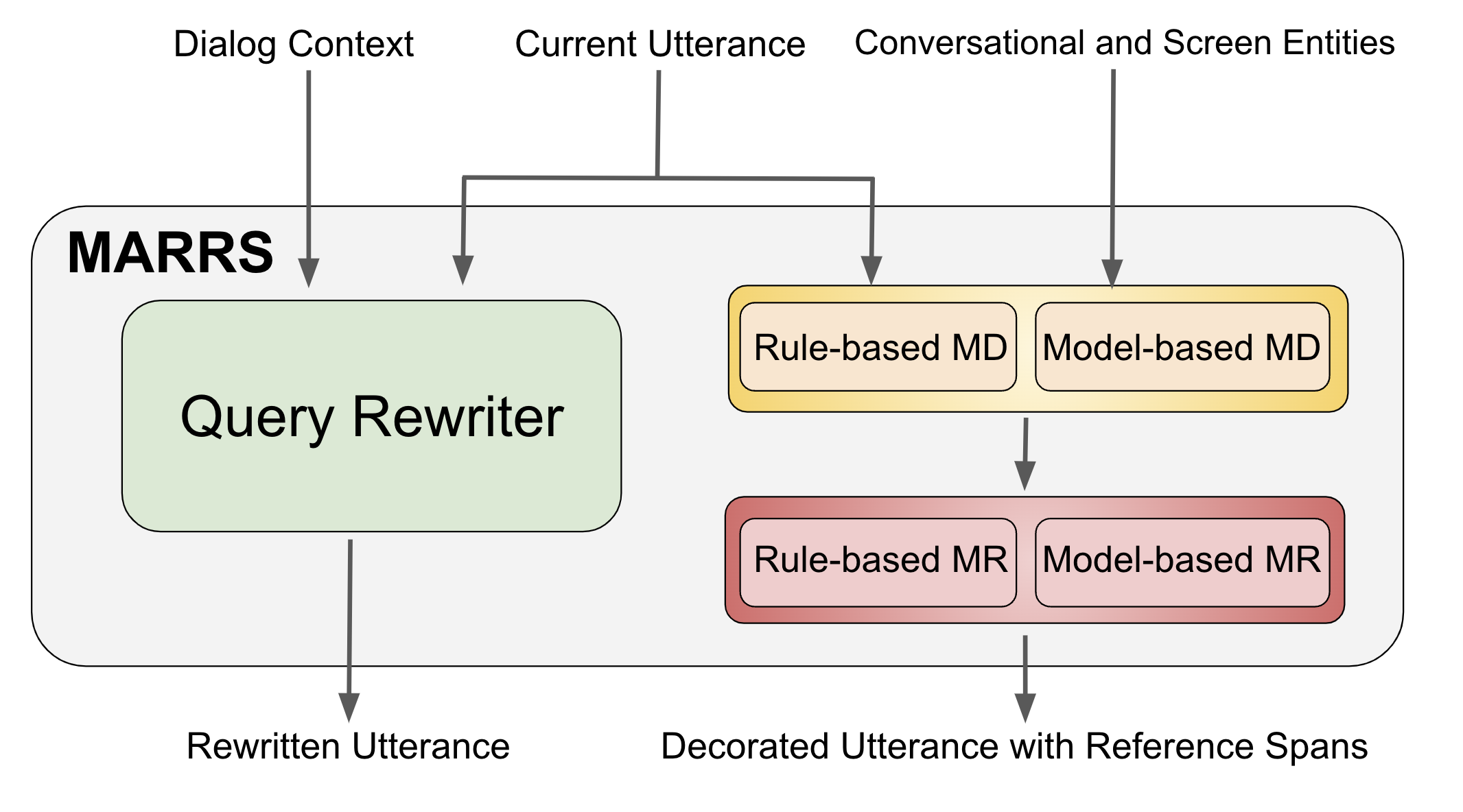}
  \caption{High level diagram show-casing how MARRS models interface with each other}
  \label{fig:marrs}
\end{figure}

\subsubsection{Model-based MD}
This predicts which sequences of tokens (spans) need to  be resolved to an entity. The model takes in token embeddings and enumerates all spans consisting of start and end token indices. For each span, the first and last token embeddings are concatenated and fed into a feed forward network that performs binary classification.  We opt for this approach instead of an LSTM or self-attention based sequence tagger because using a classifier that classifies all spans allows the model the flexibility to identify multiple (possibly overlapping) mentions, while also allowing all spans to be classified independently in parallel (as opposed to sequentially or auto-regressively). Our approach is very similar to that of \citet{lee2017end}, except that we empirically observe very little impact by removing the self-attention layer in their mention detector (primarily because our coreference dependencies tend to be much shorter than theirs), while observing very large improvements in both latency and memory. The model architecture is shown in \ref{sec:model_diag} Figure~\ref{fig:md}.

\subsubsection{Rule-based MD}

While the model detects referring expressions (which often include marker words like "this" and "that"), there are cases when the user refers to a contextual entity by name only (omitting the referring expression). In a user request like ``Call customer support'', ``customer support'' might refer to a support number on the user's screen. To keep the model light-weight, MD model does not consume entities; consequently, it is unable to detect that "customer support" is a referring expression. The Rule-based MD component bridges this gap by matching the gathered contextual entities to the utterance through smart string matching. If a contextual entity is found in the utterance, this sub-component outputs the span and the corresponding entity as a potential reference.

\subsection{Mention Resolver}

The Mention Resolver (MR) resolves references in user queries to contextual entities like phone numbers and email addresses. As with the overall system, the focus is on a low memory footprint and reusing the existing components in the pipeline. MR operates on the text and location of screen or conversational entities recognized by upstream component and the metadata of the background entities. It consumes the possible mentions identified by MD and matches each mention to zero, one or more entities, providing a relevance score for each. It includes a mixture of a rule-based system and a machine learned model. The rule-based system is high precision and extremely fast. Consequently, if it outputs a resolution, the model is not run, which yields a substantial latency reduction.

\subsubsection{Rule-based MR}
Rule-based MR utilizes a set of pre-defined rules and keywords to match references to the correct category, location and text. For example, references like ordinals are matched with regex patterns sorted by the longest match; likewise, music and movie entities can be matched by relying on the presence of verbs like `play'.

\subsubsection{Model-based MR}
We also design a modular reference resolution model, inspired by \citet{yu2018mattnet}. This is trained to score how well an entity matches with the detected mention. The entities for which score crosses a threshold are then predicted. The model contains 3 modules: 1. the category module, which matches the mention with the entity’s category; 2. the location module, which matches the mention with the entity's location; 3. the text module, which matches the mention with the text within entities, like screen texts and alarm names. Weights are computed using the request tokens to determine the aggregation of the the three module scores. 
We refer interested readers to \citet{bhargava-etal-2023-referring} for a more in-depth understanding of the model; we show the model architecture in Appendix Figure~\ref{fig:mr_model}.

\paragraph{Screen-based}
The entities on screen are the candidate referents here. Each entity has a category like phone number and address, a bounding box representing its location on the screen and associated text values. Each of the three modules thus receive input for screen entities, and play a key role in understanding diverse references.

\paragraph{Conversational}
Here, a user's previous conversational interaction and the VA's responses are considered as referents. In such cases, descriptive references made by a user, such as when referring to addresses (Eg: "Show me coffee shops near me" -> "Call the one on Homestead Road") are to be handled by the text module. The location module is critical in resolving ordinal and spatial references (Eg: "Show me coffee shops near me" -> "Call the bottom one" or "Call the last one").

\paragraph{Background}
In this case, entities relevant to background tasks are potential referents. These tasks may include user-initiated tasks, such as music that's playing in the background, or system-triggered tasks, such as a ringing alarm or a new notification. The category module is particularly important here, since a user's references tend to be related to the type of the referent (Eg: ``pause it'' likely refers to music or a movie, while ``stop that'' could also refer to an alarm or a timer).

\subsection{Query Rewriter}

The Query Rewriter (QR) is the component that rewrites the last user utterance in a conversation between the user and the VA into a context-free utterance such that it can be fully interpreted and understood without the dialog context.
Three use cases mentioned in Section~\ref{sec:intro} can be tackled through rewriting: Anaphora, Ellipses, and Corrections by Repetition.
The output rewritten utterance is provided as an alternative to downstream components together with the original utterance, to provide them with the flexibility of choice.

Again, for the sake of latency and privacy, the QR model is run on device along with MD and MR. Unlike the more complex components in MD and MR, QR is merely an LSTM-based seq2seq model with a copy mechanism \cite{gu2016incorporating}.
It takes as input both conversational context (i.e., a sequence of interactions) and the last user query, and generates the rewritten utterance. On top of the encoder, there is a classifier that consumes the input embeddings and predicts the type of use case (`Anaphora and Ellipsis`, `Correction by Repetition` or `None`). `None` means no rewriting is required, in which case, to further reduce latency, no decoder inference needs to be run, and the module can simply pass-through the input utterance as the output. This classification signal is also sent as part of the output to downstream systems for their use.
The model architecture can be referred to Figure~\ref{fig:qr_diagram} in Appendix.

\section{Datasets} \label{sec:data}

Since the system handles varied kinds of references, different datasets are used for training the different components. We briefly describe here the datasets used by the system.

For Screen Entity Resolution, we collect requests referring to entities on screens by showing screenshots containing entities to annotators. One entity is highlighted as the target entity. Annotators are asked to provide requests that refer uniquely to the marked entity. The collected requests are sent through another round of annotation for getting the mentions, in order to train MD. Interested readers can refer to \citet{bhargava-etal-2023-referring} for more details on the data collection. The requests and mentions collected alongside the entities are used to train the model-based MD and MR, as well as to evaluate the overall system.

For Conversational Entity Resolution, we show annotators a list of entities similar to the Agent turn in Figure~\ref{fig:conv_rr_example_flow}. These lists are synthetically generated based on the domain. Annotators are asked to provide a request referring to any one entity in the list (similar to the second User turn in Figure~\ref{fig:conv_rr_example_flow}), along with the the mention (to train MD) and the list index of the entity being referred (to train MR).

For Entity Resolution, we additionally have a synthetic data pipeline. Requests are generated through templates like `play [this]' or `share [that address] with John', with the marked mention used to train MD. A synthetic list of targeted entities is part of the templates, and synthetic negative entities are added while training MR. 

For Query Rewriting, we use mined data from the anonymized opt-in usage data. We first identify the opportunities where user experience can be significantly improved if the desired features are enabled.
In particular, for the use case of anaphora and ellipsis, we identify user queries discussing the same entity in two consecutive turns without the use of any referring expressions (context-free query).
We then ask annotators to simplify these complex queries to provide queries in a more natural way (context-dependent query).
For the use case of correction by repetition, we recognize queries where the user tapped on the transcribed prompt to edit the query into something else. The resulting utterance serves as a complete context-free query.
We then prepend the edit parts with common prefixes such as `\textit{I said}' to synthesize the context-dependent query.
By doing so, we simulate the pair of original queries and their rewrites for the two desired features.
More detailed statistics and examples of both use cases are shown in Appendix~\ref{app:data_stats} and \ref{app:data_collection}.

\begin{table}
    \centering
    \begin{tabular}{|c|l|c|c|} \hline 
         Model&Dataset/Task&  F1& EM\\ \hline \hline
         \multirow{2}{*}{QR}&AER&  91.44    & 87.83      \\ \cline{2-4}
         &CbR&  88.12                 & 71.44     \\ \hline \hline
          \multirow{3}{*}{MDMR}&Screen& 83.39 & 80.8 \\ \cline{2-4} 
          &Conversational & 89.85 & 91.50 \\ \cline{2-4}
          &Synthetic& 97.56 & 96.90 \\\hline
    \end{tabular}
    \caption{Experimental results for QR and MDMR. Here, the Synthetic MDMR dataset tests performance of both Conversational and Background Entity Resolution use-cases.}
    \label{tab:results}
    \vspace{-0.5\baselineskip}
\end{table}

\section{Experimental Results}

\subsection{Metrics} \label{sec:metrics}
We compute a bag of words token level F1 metric on the subset of tokens that are present in the target rewrite, but not in the corresponding context dependent query. This metric reflects the model's ability to carry over tokens from previous context. We also calculate an exact string match accuracy (EM) between the model prediction and the target rewrite as a more strict comparison. Metrics for anaphora and ellipsis resolution (AER) and correction by repetition (CbR) are measured separately. 

For reference resolution, metrics are computed by comparing the true target entities with the predicted entities (for which scores cross the threshold). Similar to above, we report F1 and exact match metrics. Here, exact match is 1 if the predicted entities over all predicted references exactly match the true target entities, and is 0 if any additional or missing predicted entities exist. 


\subsection{Performance}

We present an overview of our system performance as measured on the datasets described in Section~\ref{sec:data} in Table~\ref{tab:results}, with additional results in Appendix~\ref{app:results}. We find that our models afford excellent performance despite being extremely small, lightweight enough with respect to both model size and runtime inference latency to potentially deploy them to a low-power device. In particular, the reported results use a QR model with just 4.5M parameters with 1-layer 128-dim LSTMs as encoder and decoder; the MD and MR models are even lighter, with just 116k and 196k parameters respectively.

\section{Conclusions}
 In this paper, we propose and provide a system-level overview of MARRS, a low memory system that combines multiple models to solve context understanding. Our design choices offer an interpretable and agile system. This system can improve user experience in a multi-turn dialogue agent in a fast, efficient, on-device and privacy-preserved manner. 

\bibliography{anthology,custom}
\bibliographystyle{acl_natbib}

\newpage

\appendix

\section{Appendix}
\label{sec:appendix}
\subsection{Detailed Model Architectures}
\label{sec:model_diag}
\begin{figure}[!htbp]
  \centering
  \includegraphics[width=0.9\linewidth]{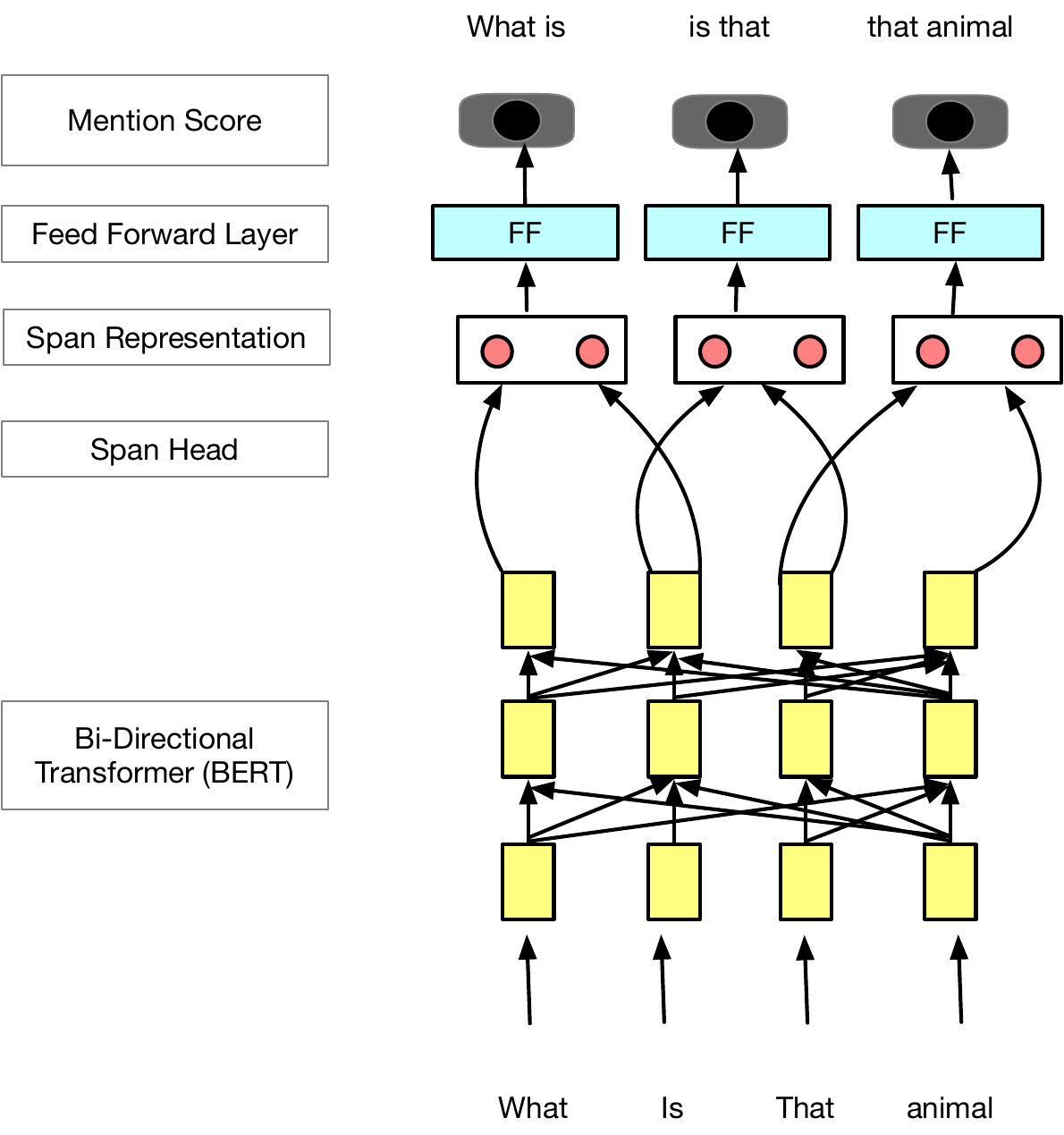}
  \caption{MD model overview}
  \label{fig:md}
\end{figure}

\begin{figure}[!htbp]
    \centering
    \includegraphics[width=1\linewidth]{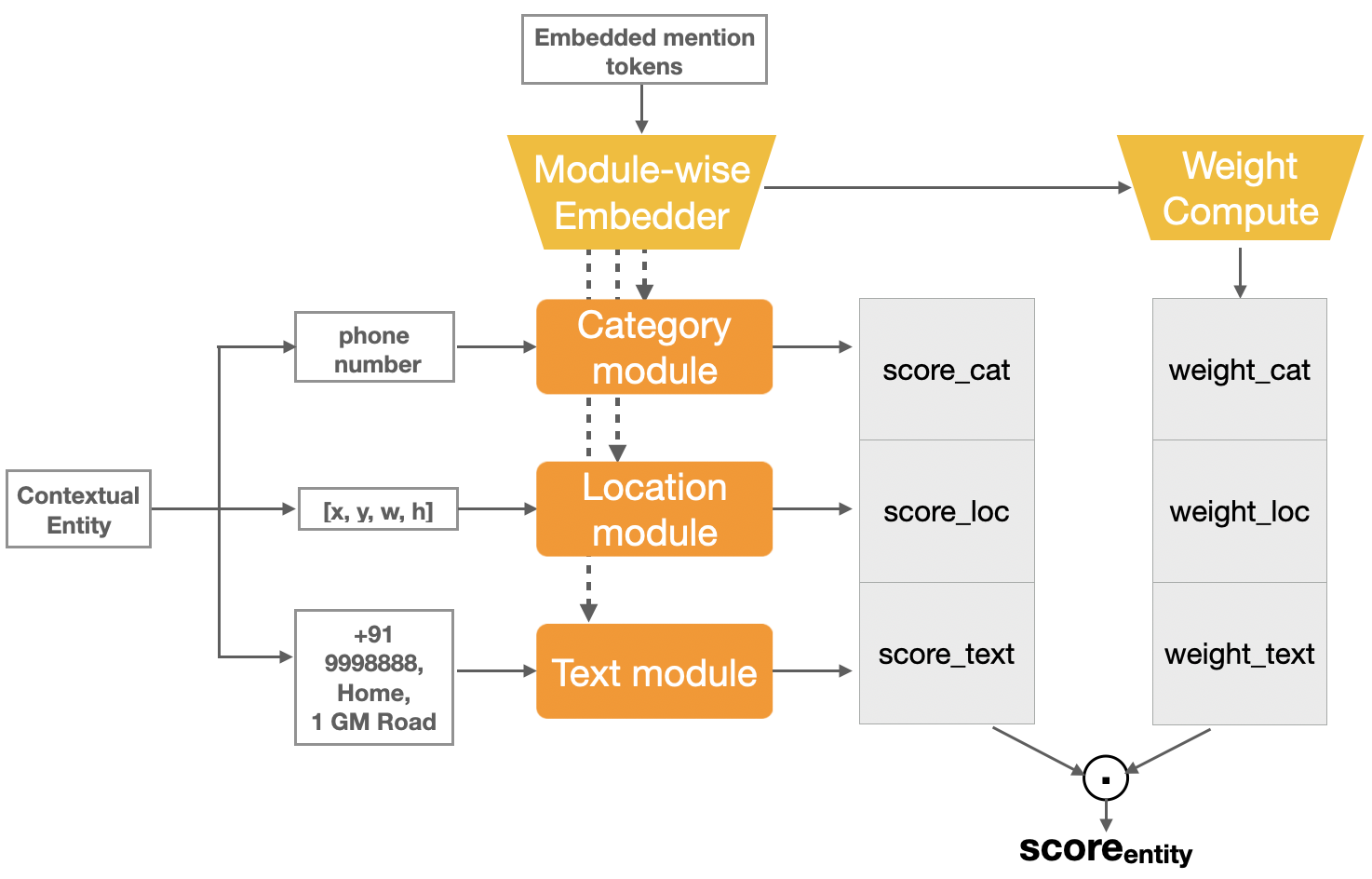}
    \caption{MR model overview}
    \label{fig:mr_model}
\end{figure}

\begin{figure}[!htbp]
  \centering
  \includegraphics[width=1\linewidth]{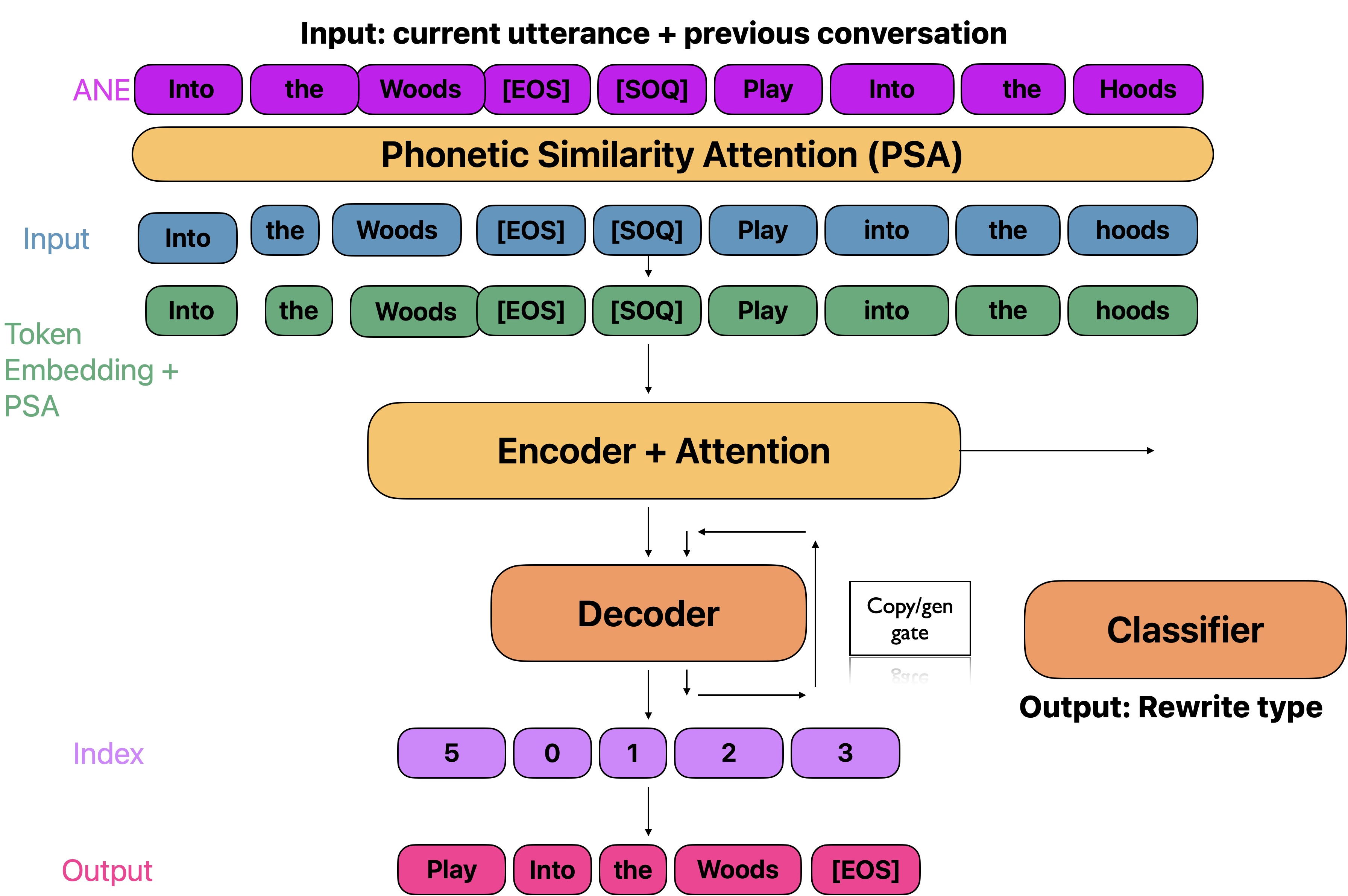}
  \caption{QR model overview}
  \label{fig:qr_diagram}
\end{figure}

This section provides the model architectures of MD (Figure~\ref{fig:md}), MR (Figure~\ref{fig:mr_model}) and QR (Figure~\ref{fig:qr_diagram}) adopted in the MARRS system. 

\subsection{Data Statistics} \label{app:data_stats}

Here, we present more detailed statistics around our datasets. In particular, we present the sizes of each dataset: we show how much data was used from each set for training, validation and testing. We present these numbers in Table~\ref{tab:data_stats}.

\begin{table}[!h]
  \centering
  \vspace{0.5\baselineskip}
  \begin{tabular}{|@{\hskip4pt}c@{\hskip3.5pt}|l@{\hskip3.5pt}|@{\hskip4pt}c@{\hskip4pt}|@{\hskip3pt}c@{\hskip3pt}|@{\hskip3pt}c@{\hskip3pt}|} \hline 
       Model&Dataset/Task&  Train & Val & Test \\ \hline \hline
        \multirow{3}{*}{MDMR}&Screen& 7.3k & 0.7k & 1.9k \\ \cline{2-5} 
        &Conversational & 2.3k & 0.4k & 1.2k \\ \cline{2-5}
        &Synthetic& 3.9k & 0.5k & 1.1k \\\hline \hline
       \multirow{2}{*}{QR}&AER&  300.3k             & 37.3k   &   37.2k     \\ \cline{2-5}
       &CbR&  317.5k                 &  39.7k &  39.7k  \\ \hline
  \end{tabular}
  \caption{Dataset sizes used for the MDMR and QR models.}
  \label{tab:data_stats}
\end{table}

\subsection{Data Collection} \label{app:data_collection}
\begin{figure}[htbp]
\vspace{0.5\baselineskip}
    \centering
    \includegraphics[width=1\linewidth]{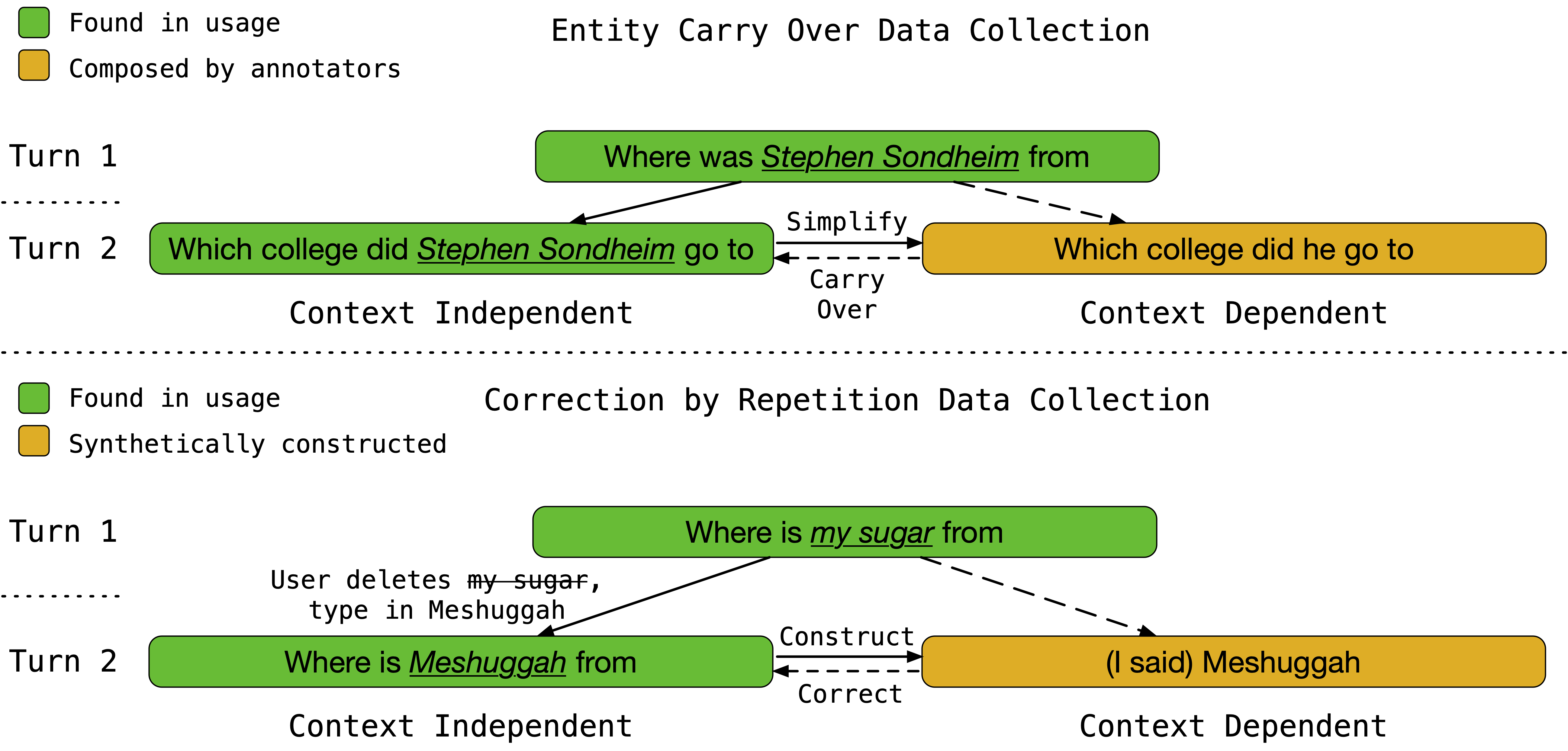}
    \caption{Illustration of data collection process for Anaphora, Ellipses and Correction by Repetition. Examples shown are author-created examples based on anonymized and randomly sampled virtual assistant logs. In both examples, utterances in green are improvement opportunities found in real-world usage, utterances in yellow are either annotated or synthetically generated. During data collection phase, we follow the solid lines. During model training, we follow the dotted lines.}
    \label{fig:data}
\end{figure}

This section provides an example of anaphora, ellipsis and correction by repetition of our data mining methods, as shown in Figure \ref{fig:data}.

\subsection{Results} \label{app:results}

\begin{table}[!t]
  \centering
  \begin{tabular}{|@{\hskip4pt}c@{\hskip3.5pt}|l@{\hskip5pt}|@{\hskip4pt}c@{\hskip4pt}|@{\hskip4pt}c@{\hskip4pt}|@{\hskip4pt}c@{\hskip4pt}|} \hline 
       Model&Dataset/Task&  P & R & F1 \\ \hline \hline
        \multirow{3}{*}{MD}&Screen & 89.66 & 95.74 & 92.60  \\ \cline{2-5} 
        &Conversational  & 85.30 & 92.60 & 88.80  \\ \cline{2-5}
        &Synthetic & 99.00 &  99.70 & 99.30 \\\hline \hline
        \multirow{3}{*}{MR}&Screen & 87.99 & 85.87 & 86.92 \\ \cline{2-5} 
        &Conversational & 85.62 & 96.91 & 90.92  \\ \cline{2-5}
        &Synthetic & 98.09 & 97.53 & 97.81 \\\hline \hline
        \multirow{3}{*}{MDMR}&Screen& 86.85 & 80.20 & 83.39 \\ \cline{2-5} 
        &Conversational & 84.70 & 95.66 & 89.85 \\ \cline{2-5}
        &Synthetic& 97.92 & 97.21 & 97.56 \\\hline \hline
       \multirow{2}{*}{QR}&AER&  92.48             & 90.42   &   91.44     \\ \cline{2-5}
       &CbR&  93.31                 & 83.48  &  88.12  \\ \hline
  \end{tabular}
  \caption{Precision, recall and F1 scores for the MD, MR and QR models.}
  \label{tab:pr_results}
  \vspace{-0.5\baselineskip}
\end{table}

In this section, we present a deep dive of the results shown in Section~\ref{sec:metrics}. In particular, we present detailed precision, recall and F1 numbers for our QR, MD and MR models, as well as our joint MDMR performance. 
Note that in the case of QR, following \citet{quan-etal-2019-gecor}, we measure the F1 score by comparing generated rewrites and references for only the rewritten part of user utterances. This highlights the model's ability to carry over essential information through rewriting. In case of MR, we consider the ground truth mentions and compute the metrics by comparing the predicted entities with the true target entities. For MDMR, we use the predicted mentioned from MD to run MR, and then compute metrics by comparing all the predicted entities with the true entities. 

\end{document}